\documentclass[runningheads]{llncs}

\usepackage{pdfsync,amssymb,amsmath,ifpdf,color}

\newcommand{\R}{{\mathbb{R}}} 
\newcommand{\T}{{^\text{T}}}             % for matrix transpose 
\newcommand{\W}[1]{\mathbf{W}^\text{#1}} % matrix
 
\newcommand{\n}[1]{{N_\text{#1}}} 
\def\u{\mathbf{u}} % steal the \u accent
\newcommand{\x}{\mathbf{x}}
\newcommand{\Dx}{\mathbf{\tilde{x}}}
\newcommand{\teach}{\text{target}}
\newcommand{\y}{\mathbf{y}} 
\newcommand{\yt}{\mathbf{y}^\teach} 
\newcommand{\concat}[2]{{[#1;#2]}}
\newcommand{\X}{\mathbf{X}}
\newcommand{\Y}{\mathbf{Y}}
\newcommand{\Yt}{\mathbf{Y}^\teach}
\newcommand{\U}{\mathbf{U}}
\newcommand{\I}{\mathbf{I}}

\newcommand{\OO}{{\BigO}}

\usepackage{booktabs}% More proffesional look of tables
\newcommand{\tableHeader}[1]{\textbf{#1}}

\usepackage{float}
\usepackage{multirow}
\usepackage{graphicx}

\usepackage[hang]{subfigure}
\usepackage{adjustbox} % resize table to fit the page 
\usepackage{tabularx}
\usepackage{mdwtab} % persistent lines
\newcommand{\BigO}{\mathcal{O}}

\definecolor{refcolor}{rgb}{0,0,0.5}
\ifpdf
\PassOptionsToPackage{hyphens}{url}
\usepackage[% 
  pdftitle={Efficient cross-validation of echo state networks},% 
  pdfauthor={Mantas Lukosevicius,Arnas Uselis},% 
  pdfkeywords={Echo State Network, Reservoir Computing, Recurrent Neural Networks, Cross-Validation, Time-Complexity},% 
  pdfstartview=FitV,% 
  bookmarks=true,% 
  bookmarksopen=true,% 
  breaklinks=true,% 
  colorlinks=true,% 
  linkcolor=refcolor,%
  anchorcolor=blue,% 
  citecolor=refcolor,%
  filecolor=blue,% 
  menucolor=blue,% 
  urlcolor=refcolor,%
  pdfpagelabels]{hyperref} 
\else 
\usepackage[% 
  breaklinks=true,% 
  colorlinks=true,% 
  linkcolor=blue,%
  anchorcolor=blue,% 
  citecolor=blue,%
  filecolor=blue,% 
  menucolor=blue,% 
  urlcolor=blue]{hyperref} 
\fi 

\begin{document}
\title{Efficient Cross-Validation of Echo State Networks}
\author{Mantas Luko\v{s}evi\v{c}ius\inst{1}\orcidID{0000-0001-7963-285X} \and
Arnas Uselis\inst{1}}
\authorrunning{Luko\v{s}evi\v{c}ius and Uselis}
\institute{
Kaunas University of Technology, Studentu st. 50-–406, 51368 Kaunas, Lithuania
\email{\{mantas.lukosevicius,arnas.uselis\}@ktu.edu}}
\maketitle              % typeset the header of the contribution
\begin{abstract}

Echo State Networks (ESNs) are known for their fast and precise one-shot learning of time series. But they often need good hyper-parameter tuning for best performance. For this good validation is key, but usually, a single validation split is used. In this rather practical contribution we suggest several schemes for cross-validating ESNs and introduce an efficient algorithm for implementing them. The component that dominates the time complexity of the already quite fast ESN training remains constant (does not scale up with $k$) in our proposed method of doing $k$-fold cross-validation. The component that does scale linearly with $k$ starts dominating only in some not very common situations. Thus in many situations $k$-fold cross-validation of ESNs can be done for virtually the same time complexity as a simple single split validation. Space complexity can also remain the same. We also discuss when the proposed validation schemes for ESNs could be beneficial and empirically investigate them on several different real-world datasets. 

%The abstract should briefly summarize the contents of the paper in 150--250 words.

\end{abstract}

\section{Introduction}

Echo State Network (ESN) \cite{Jaeger01a,JaegerHaas04,Jaeger07scholarpedia} is a recurrent neural network training technique of reservoir computing type \cite{LukoseviciusJaeger09}, known for its fast and precise one-shot learning of time series. But it often needs good hyper-parameter tuning to get the best performance. For this fast and representative validation is very important. 

Validation aims to estimate how well the trained model will perform in testing. Typically ESNs are validated on a single data subset. This single training-validation split is just one shot at estimating the test performance. Doing several splits and averaging the results can make a better estimation. This is known as cross-validation (\cite{Stone74} is often cited as one of the earliest descriptions), as the same data can be used for training in one split and for validation in another and vice versa. 

% https://stats.stackexchange.com/questions/26696/who-invented-k-fold-cross-validation

In this contribution we suggest several schemes for cross-validating ESNs and introduce an efficient algorithm for implementing them. We also test the validation schemes on 
%TODO
five different real-world datasets.

The goal of the experiments here is not to obtain the best possible performance on the datasets, but to compare different validation methods of ESNs on equal terms. The best performance here is often sacrificed for the simpler models and procedures. Therefore we use classical ESNs here, but the proposed validation schemes apply to any type of reservoirs with the same time and space complexity savings, as long as they have the same linear readouts. 

We introduce our ESN model, training, and notation in Section \ref{esnBasics}, discuss different classes of tasks that might be important for validation in Section \ref{tasks}, discuss cross-validation nuances of time series in Section \ref{temporalXVal}, suggest several cross-validation schemes for ESNs in Section \ref{validations}, suggest several ways of producing the final trained model in Section \ref{finalModel}, and introduce a time- and space-efficient algorithm for cross-validating ESNs in Section \ref{algorithm}. We also report empirical experiments with different types of data in Section \ref{experiments} and conclude with a discussion in Section \ref{discussion}.

%\section{Previous Work?}

%Many practical aspects of using ESNs have been discussed in \cite{Lukosevicius12a}. 
\subsection{Basic ESN Training}\label{esnBasics}

Here we introduce our ESN model and notation following \cite{Lukosevicius12a}. 

The typical update equations of ESN are
\begin{equation}
\Dx(n)= \tanh \left( \W{in}\concat{1}{\u(n)}+\W{} \x(n-1) \right),
\label{eq:esnupdate} 
\end{equation} 
\begin{equation} 
\x(n)=(1-\alpha)\x(n-1)+\alpha \Dx(n),
\label{eq:leakyupdate} 
\end{equation} 
where $\x(n) \in \R^\n{x}$ is a vector of reservoir neuron activations and $\Dx(n) \in \R^\n{x}$ is its update, all at time step $n$, $\tanh(\cdot)$ is applied element-wise, $\concat{\cdot}{\cdot}$ stands for a vertical vector (or matrix) concatenation, $\W{in}\in \R^{\n{x} \times (1+\n{u})}$ and $\W{} \in \R^{\n{x} \times \n{x}}$ are the input and recurrent weight matrices respectively, and $\alpha \in (0,1]$ is the leaking rate. 

The linear readout layer is typically defined as
\begin{equation}
\y(n)=\W{out}[1;\u(n);\x(n)],
\label{eq:linreadout}
\end{equation}
where $\y(n) \in \R^\n{y}$ is the network output and $\W{out} \in \R^{\n{y}\times\n{r}}$ the output weight matrix. We denote $\n{r} = 1+\n{u}+\n{x}$ as the size of the ``expanded'' reservoir $[1;\u(n);\x(n)]$ for brevity.

Equation \eqref{eq:linreadout} can be written in a matrix notation as
\begin{equation} 
\Y = \W{out}\X, 
\label{eq:readoutM} 
\end{equation}
where $\Y \in \R^{\n{y} \times T}$ are all $\y(n)$ and $\X \in \R^{\n{r} \times T}$ are all $[1;\u(n);\x(n)]$ produced by presenting the reservoir with $\u(n)$, both collected into respective matrices by concatenating the column-vectors horizontally over the training period $n=1,\ldots,T$. We use here a single $\X$ instead of $[\mathbf{1};\U;\X]$ for notational brevity. 

%(Feedback connections $\W{fb}$ from $\y(n-1)$ to $\Dx(n)$ in \eqref{eq:esnupdate}.)

Finding the optimal weights $\W{out}$ that minimize the squared error between $\y(n)$ and $\yt(n)$ amounts to solving a system of linear equations 
\begin{equation} 
\Yt = \W{out}\X, 
\label{eq:regressTargetM} 
\end{equation}
where $\Yt \in \R^{\n{y} \times T}$ are all $\yt(n)$, with respect to $\W{out}$ in a least-square sense, i.e., a case of linear regression. In this context $\X$ can be called the \emph{design matrix}. The system is typically overdetermined because $T \gg \n{r}$.

The most commonly used solution to \eqref{eq:regressTargetM} in this context is ridge regression:
\begin{equation} 
\W{out}=\Yt \X^\T \left(\X\X^\T + \beta \I \right)^{-1},
\label{eq:ridgeRegress} 
\end{equation} 
where $\beta$ is a regularization coefficient and $\I$ is the identity matrix. %There is no need to rerun the model through the data with every value $\beta$, because none of the other variables in \eqref{eq:ridgeRegress} are affected by its changes. 
It is advisable to set the first element of $\I$ to zero to exclude the bias connection from the regularization. 

For more details on generating and training ESNs see \cite{Lukosevicius12a}.

\section{Validation in Echo State Networks}\label{proposedMethods}

In this chapter we discuss the validation options for ESNs, and propose an efficient algorithm for cross-validation.

%TODO rewrite:
%
%In this chapter we will show that ESNs allow for efficient implementation of training and validation, especially cross-validation, because we often do not need to rerun the reservoir with the training data. 
%
%This feature of ESNs is particularly helpful when searching for good meta-parameters. 

%(Time can have importance)

\subsection{Different Tasks}\label{tasks}

Some details of implementation and computational savings depend on what type of task we are learning. Let us distinguish three types of temporal machine learning tasks:
\begin{enumerate} 
    \item \textbf{Generative} tasks, where the computed output $\y(n)$ comes back as (part of) the input $\u(n+k)$. This is often pattern generation or multi-step time series prediction in a generative mode.\footnote{Note that this can alternatively be implemented with feedback connections $\W{fb}$ from $\y(n-1)$ to $\Dx(n)$ in \eqref{eq:esnupdate}.}
    \item \textbf{Output} tasks, where the computed output time series $\y(n)$ does not come back as part of input. This is often detection or recognition in time series, or deducing a signal from other contemporary signals. 
    \item \textbf{Classification} tasks, of separate (pre-cut) finite sequences, where a class $\y$ is assigned to each sequence $\u(n)$. 
\end{enumerate}

For the latter type of tasks we usually store only an averaged or a fixed number of states $\x(n)$ for every sequence in the state matrix $\X$. It is similar to a non-temporal classification task.

The experiments Section \ref{experiments} is structured according to this distinction. % with tasks of every type.

\subsection{Cross-Validation in Time Series}\label{temporalXVal}

%Cross-validation of static data. 

$k$-fold cross-validation, arguably the most popular cross-validation type, is a standard technique in static (non-temporal) machine learning tasks where data points are independent of each other. Here the data are partitioned into $k$ usually equal folds, and $k$ different train-validate splits of the data are done, where one (each time different) fold is used for validation and the rest for testing.

Temporal data, on the other hand, are time series or signals, often a single continuous one. They are position-dependent. Cross-validation in them is a bit less intuitive and popular.

A classical option for ESNs is the static split of the data into initialization, training, validation, and testing, in that order in time. The short initialization (also called transient) phase is used to get the state of the reservoir $\x(n)$ ``in tune'' with the input $\u(n)$ \cite{Lukosevicius12a}. This sequence is finite, and often quite short, because ESNs possess the echo state property \cite{YildizEtAl12}. Initialization is only necessary before the first (training) phase, because the subsequent phases can take the last $\x(n)$ from the previous phase if data continue without gaps. In generative tasks the real (future) outputs $\y(n)$ are substituted with targets $\yt(n)$ in inputs, known as ``teacher forcing'', to break the cyclic dependency. 

Because the memory of ESN is preserved in its collected state, and the classical output that is learned is memory-less, we can do the instant switches between phases. Exploiting this same Markovian property, cross-validation with ESNs is rather straightforward. In other temporal models this can be more involved \cite{BergmeirEtAl18}.

% (\url{https://robjhyndman.com/hyndsight/tscv/} \cite{BergmeirEtAl18}) Cross-validation for time series has been previously discussed in detail in \cite{BergmeirEtAl18}.

%TODO: why cross-validation makes sense.

We see the following intuitive cases when using cross-validation on time series could be beneficial:
\begin{itemize}
    \item When the data are scarce, cross-validation efficiently uses all the available data for both training and validation.
    \item Combining the models trained on different folds could be a form of (additional) regularization, improving stability. 
    \item When the process generating the data are slightly non-stationary and it ``wanders'' around, cross-validation increases the chances that the testing interval is adequately covered by the model. However, if it ``drifts'' in one direction validating and tuning the hyper-parameters on the data interval adjacent to the testing one (i.e., the classic validation) might be the best option.
\end{itemize}
In particular, we do not expect cross-validation to be beneficial on stationary synthetic long time series, like chaotic attractors, as it does not matter which (and to some extent how much if the data are ample) sections are taken for training or validation.

\subsection{Validation Schemes}\label{validations}

Here we suggest several validation schemes for ESNs.

\begin{figure}%
    \centering
    \includegraphics[width=.75\textwidth]{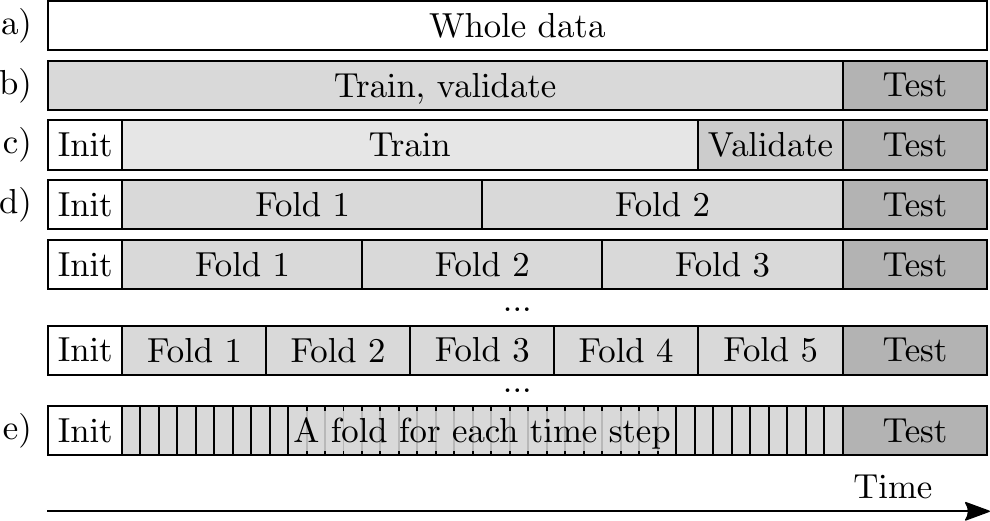} 
        \caption{Splitting the data: a) all the available data; b) splitting-off the testing set; c) a static classical (SV) initialization, training, and validation split for ESNs; d) splitting data into folds 2 and up for n-fold cross-validation; e) the maximum amount of folds for leave-one-out cross-validation.}
    \label{fig:data_splits}%
\end{figure}

We firstly split the testing part off the end which is independent of the validation scheme and is left for testing it as illustrated in Figure \ref{fig:data_splits} b). The classical ESN validation scheme explained above is presented in c). We will refer to it as \textbf{static validation} (SV). Here we also investigate alternative validation schemes where the data left from initialization and testing are used for training and validation differently and iteratively. 

To investigate the $k$-fold cross-validation of ESNs we split the data into $k$-folds. The number of folds $k$ can be varied from 2, as in Figure \ref{fig:data_splits} d), up to available data/time points ending up with leave-one-out cross-validation e).

%(If several time series and all are split into testing, all series should be split like this in parallel?)

In addition to the classical SV split we investigate these validation schemes of using the data between the initialization and testing parts:
\begin{enumerate}

    \item \textbf{$k$-fold cross-validation} (CV). The data are split into $k$ equal folds. Training and validation are performed $k$ times, each time taking a different single fold for validation and all the rest for training.  

    \item \textbf{$k$-fold accumulative validation} (AV). First, we split the ``minimum'' required amount for training only off the beginning, then we divide the rest of the data into $k$ equal folds. Training and validation are performed $k$ times, each time validating on a different fold, similarly to CV, but only training on all the data preceding the validation fold. 
    
  \item \textbf{$k$-fold walk forward validation} (FV) is similar to AV: the splitting is identical and validation is done on the same folds, but the training is each time done only on the same fixed ``minimal'' amount of data directly preceding the validation fold. 

\end{enumerate}

\begin{figure}%
    \centering
    \includegraphics[width=\textwidth]{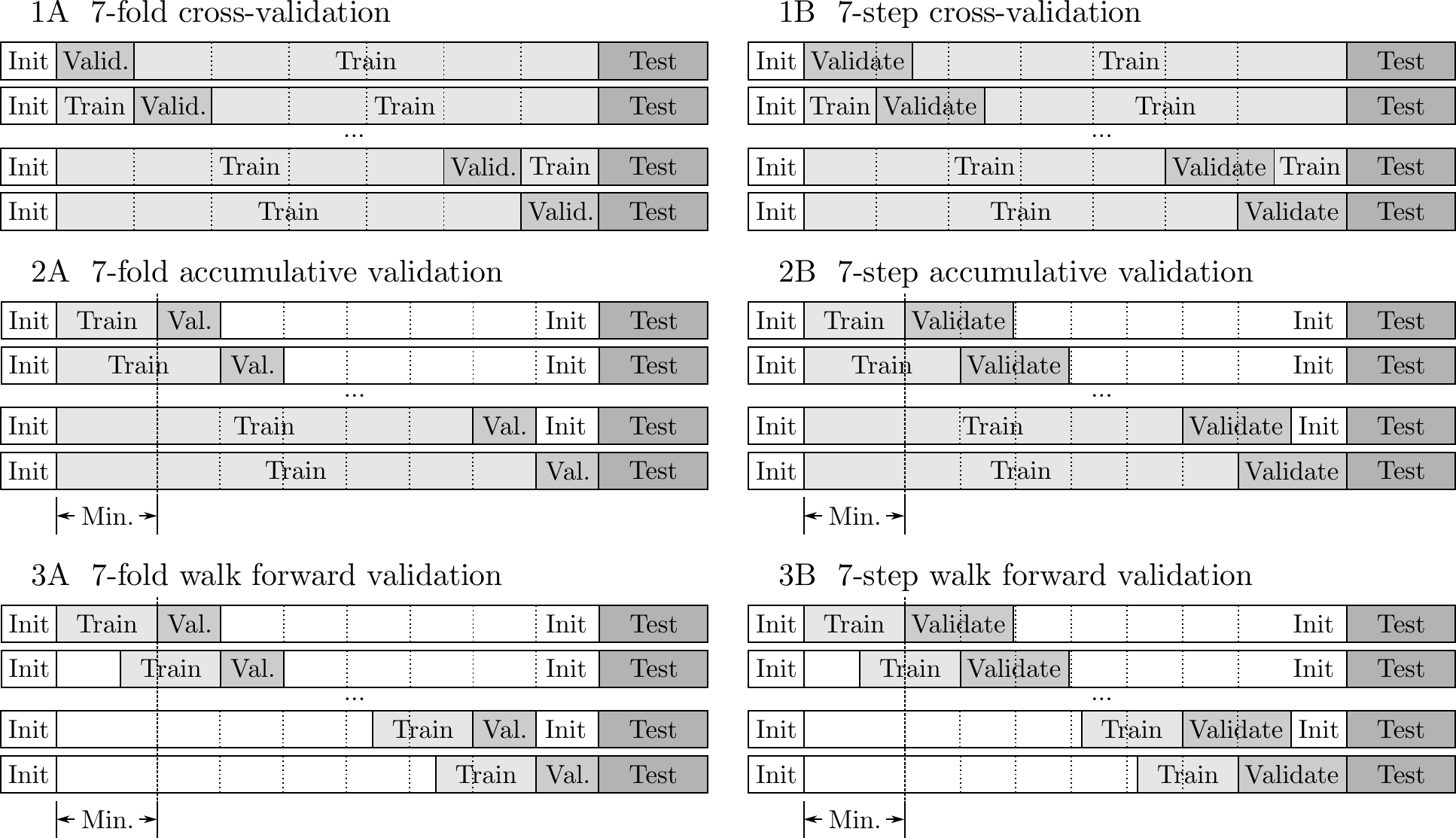} 
        \caption{Different validation schemes used.}
    \label{fig:validation_types}%
\end{figure}

The three validation schemes are illustrated in the left column of Figure \ref{fig:validation_types}. 

Each validation scheme has some rationale behind it. 

In CV all the data are used for either training or validation in each split. Also, all the data have been used exactly $k-1$ times for training and 1 time for validation. This has its benefits explained in Section \ref{finalModel}. What is a bit unusual for temporal data here is that training data also come later in the sequence than the validation data.
But this should not necessarily be considered a problem.
%But this should not be a problem if we assume the process generating the data are stationary (and would not be able to learn much otherwise). 
In fact, for time series output and classification tasks the columns of $\X$ and $\Yt$ could in principle be randomly shuffled, before applying $k$-fold cross-validation, as is common in non-temporal tasks. CV is geared towards training the model once on a fixed well-representative dataset.

AV emulates the classical static training and validation SV $k$ times by each time training on all the available data that come before the validation fold and then validating. This scenario would also happen when the model is repeatedly updated with newly available data. Here we do not allow our model to ``peek into the future''. The downsides are that models are not trained on the same amount of data and more models are trained on the beginning of the data than the ending. 

FV is similar to AV but keeps the training data length constant like CV, which is more consistent when selecting good hyper-parameters. Training length can be set to match several folds, which is quite common when doing walk forward validation. Here the assumption of stationarity of generating process is weaker, models are trained and validated ``locally'' in time.

CV, AV, and FV all use progressively less data for training thus, in general, are progressively less advisable when data are scarce. 

We have also investigated counterparts for the three validation schemes with validation ``folds'' set longer, independent of $k$. They are illustrated in the right column of Figure \ref{fig:validation_types}. We call them ``$k$-step'' as opposed to ``$k$-fold''. We are inventing terminology here when we could not find (a consistent) one in the literature. 

These ``$k$-step'' validation schemes are mostly relevant to realistically validate generative tasks, where errors tend to escalate over time when running in the generative mode. For other types of tasks, validation length is usually not important when we compute time-averaged errors. The validation length here can be set to match a testing length in a realistic use scenario. It can also be set to match the length of several steps, which would use data for training and validation more consistently.

%(optionally validation can be several consecutive folds.)
%(overlapping bigger validation ``folds''  validation ``fold'' can have the same length as testing (16\%).)
%Split: 4\% init at the beginning, 16\% testing at the end. if we take 5-fold x-val, folds are also 16\%.

%Validation on training?..

\subsection{Final Trained Model}\label{finalModel}

Validation results are usually averaged over splits for every hyper-parameter set and the best set is selected. We investigate several ways of producing the final trained model with the best hyper-parameters for testing:
\begin{itemize}
    \item \textbf{Retrain} ESN on all the training and validation data;
    \item \textbf{Average} $\W{out}$s of ESNs that have been trained on the $k$ splits;
    \item Select the ESN that validated \textbf{best} among the $k$.
    % take the fold that has been trained on the most recent data?
\end{itemize}

Each method again has its own rationale. Retraining uses all the available data to an ESN in a straightforward way. However, this requires some additional computation and the hyper-parameters might no longer be optimal for the longer training sequence. Averaging $\W{out}$s introduces additional regularization, which adds to the stability of outputs. This method might make less sense on AV, since models are trained on different amounts of data. The best-validated split, on the other hand, was likely trained on the hardest parts of the data (and validated on the easiest). A weighting scheme among splits could also be introduced when combining ESNs, as well as time step weighting as discussed in \cite{Lukosevicius12a}. 

\textbf{Regularization} parameter is grid-searched for every split individually, as opposed to other hyper-parameters, that are searched in outside loops (see \cite{Lukosevicius12a} for why this is efficient). For average and best ESNs with their best regularization for that particular fold can be used, whereas for retraining the regularization that is best on average (over the folds) is utilized. Regularization is most important in generative tasks.

For the classic static split SV we also have two options: to either retrain the model on the whole data or to use the validated model as it is. 

%Averaged over folds -- adds regularization a bit like dropout. Bias for end folds?

%Best fold (for comparison?)

%\subsection{Validation and Regularization}

%Memory permitting, there is also no need to rerun the model with the (small) validation dataset, if you can store $\X$ for it and compute the validation output by  \eqref{eq:readoutM}. This makes testing $\beta$ values computationally much less expensive than testing the other reservoir parameters.

%Ternary search -- for each fold \arnas{VS globaliai parinktas reg. degree. Cia maziau sietusi su "dropout" reguliarizacija}

%(Do not regularize bias)

%Averaging is a form of regularization.

%Either global regularization parameter, or individual for each fold (then average Wouts or retrain with average reg. parameter (not tried)). 

%Reg is most important for generative tasks. 

\subsection{Efficient Implementations}\label{algorithm}

Running the ESN reservoir \eqref{eq:esnupdate} is dominated by $\W{}\x$ which takes $\OO(\n{x}^2)$ operations per time step with dense $\W{}$, or $\OO(\n{x}^2 T)$ for all the data. This can be pushed down to $\OO(\n{x} T)$ with sparse $\W{}$ \cite{Lukosevicius12a} which is the same $\OO(\n{r} T)$ required for collecting $\X$. For computing $\W{out}$ \eqref{eq:ridgeRegress}, collecting $\X\X^\T$ takes $\OO(\n{r}^2 T)$ and the matrix inversion takes $\OO(\n{r}^3)$ operations in practical implementations. Thus the whole training of ESN (dominated by collecting $\X\X^\T$) is back to
\begin{equation}
\OO(\n{r}^2 T). 
\label{eq:Oesn} 
\end{equation}
The same \eqref{eq:Oesn} applies to training and validating ESN, as validation itself has the same complexity as simply running it. And doing a straightforward ESN $k$-fold cross-validation is 
\begin{equation}
\OO(k \n{r}^2 T).
\label{eq:Onaive} 
\end{equation} 

Space complexity can be pushed from $\OO(\n{r} T)$ for $\X$ down to $\OO(\n{r}^2)$ when collecting $\X\X^\T$ and $\Yt \X^\T$ on the fly, which also allows ESNs to be one-shot-trained on virtually infinite time sequences \cite{Lukosevicius12a}.

%Notice that the dimensions of the matrices $(\Yt \X^\T) \in \R^{\n{y} \times \n{r}}$ and $(\X\X^\T) \in \R^{\n{x} \times \n{r}}$ do not depend on the length $T$ of the training sequence in their sizes. The two matrices can be updated by simply adding the corresponding results from the newly incoming data. This one-shot training approach in principle works with an unlimited amount of data -- neither complexity of working memory nor time of the training procedure \eqref{eq:ridgeRegress} itself depend on the length of data $T$.

However, we do not need to rerun the ESN for every split. We can collect and store the matrices $\X\X^\T$ and $\Yt \X^\T$ for the whole sequence once. Then in every split we only run the reservoir on the validation fold. Validation folds should be arranged consecutively like in Figure \ref{fig:validation_types}.1A, so that after running one validation fold we can save the reservoir state $\x(n)$ for the next validation fold of the next split. We collect $\X\X^\T$ and $\Yt \X^\T$ on the validation fold and subtract them from the global ones to compute $\W{out}$ for the particular split. If we are doing output or classification task (see Section \ref{tasks}) and we can afford to store $\X$ of the validation fold in memory, we can reuse it to compute the validation output $\y(n)$ \eqref{eq:linreadout}. If not, we need to rerun the validation fold one more time for this. \textit{This way the ESN is rerun through the whole data only two or three times irrespective of $k$.}

Notice also, that the space complexity of such implementation remains $\OO(\n{r}^2)$. We could alternatively also store $\X\X^\T$ and $\Yt \X^\T$ for every fold and save one running through the data this way, by having space complexity $\OO(k\n{r}^2)$. 

The proposed method pushes down the time complexity of preparation of $\X\X^\T$'s in $k$-means cross-validation from $\OO(k \n{r}^2 T)$ which dominates in \eqref{eq:Onaive}, to $\OO(\n{r}^2 T)$. Adding the matrix inversions \eqref{eq:ridgeRegress} which are now not necessarily dominated, the propsed more efficient implementation of ESN $k$-means cross-validation has time complexity
\begin{equation}
\OO(\n{r}^2 T + k \n{r}^3).
\label{eq:Ogood} 
\end{equation} 

Thus we get a $k$ or $T/\n{r}$ time complexity speedup in a more efficient implementation \eqref{eq:Ogood} compared to naive \eqref{eq:Onaive}, depending on which multiplier is smaller. 

When the data sample length $T$ is many times larger than the ESN size $\n{r}$ (a typical case and a one where optimization is most relevant) and thus $k$ such that $k<T/\n{r}$, we can say that \textit{the proposed efficient implementation permits doing ESN $k$-folds cross-validation with the same time complexity as a simple one-shot validation. The space complexity can also remain the same.}

%\arnas{Verta pamineti, kad jeigu treniravimo seka yra labai ilga ir storinam tik $X$ ir $X^T$ bei $YX^T$ visai sekai, tai apsimoka issaugoti x steitus tik tokiose vietose, nuo kuriu generuosim validation folda. Jam sugeneravus $\X$, galima bus atimti ta X is viso $X*X^T$.} 

We have outlined an efficient method for ESN $k$-folds cross-validation (CV), but it can easily be adapted to other types of validation schemes described in Section \ref{validations}.

\section{Experiments}\label{experiments}

Having established that different validation schemes for ESNs are possible and can be implemented quite efficiently, in this section we test them empirically on several different time series datasets.

As mentioned before, the goal here is not to obtain the best possible performance but to compare different validation methods of simple ESNs on equal terms. 

%Prie kiekvieno eksperimento: dataset (ref., url, skaiciai), data split skaiciai, implementation details, grid search skaiciai, best parameters, results.

%Gerai yra lenteles, kas bendra nereikia kartoti

%Plotai - idomiausi, kiek tilps.

\subsection{Generative Mode}\label{generativeExperiments}

%Multiple variants of validation techniques were examined: 
%\begin{enumerate}
    %\item Forward Cross-Validation, while computing $W^{out}$ by:
        %\begin{enumerate}
            %\item Averaging $W^{out}$ among folds \arnas{doesn't make sense, bet gerai koreliavo valid/test kazkodel..}
            %\item Taking $W^{out}_b$, where $b$ corresponds to $b$-th fold, so that the validation error of $b$-th fold was the smallest. 
        %\end{enumerate}
    %\item (Fast) Cross-Validation, while $W^{out}$ computed by:
        %\begin{enumerate}
            %\item Averaging $W^{out}$ among folds, while each fold has its regularization degree chosen as to minimize that fold's error.
            %\item Averaging $W^{out}$ among folds, while regularization degree is the same for all of the folds.
        %\end{enumerate}
    %\item Out of Sample (SV) data split, where the last 10\% of training data are used as validation. For testing, $W^{out}$ was computed by:
        %\begin{enumerate}
            %\item Driving ESN by input of validation set. 
            %\item Retraining ESN on whole training and validation data.
        %\end{enumerate}
%\end{enumerate}
%US finished motor gasoline product supplied - generative
%Unemployment rate in US prediction, (+airline? +ozone)

%
%\begin{figure}%
    %\centering
    %\subfloat[label 1]{{\includegraphics[width=5cm]{../kodas/figures/best_pred_FCV_gasoline_False_widening.pdf} }}%
    %\qquad
    %\subfloat[pathetic]{{\includegraphics[width=5cm]{../kodas/figures/best_pred_FCV_unempl_True_OOS.pdf} }}%
    %\label{fig:example}%
%\end{figure}
 
We evaluate the proposed validation methods by examining multiple univariate datasets of increasing sizes:
\begin{itemize}
    \item \textbf{Labour}: ``Monthly unemployment rate in US from 1948 to 1977'' dataset\footnote{Publicly available at \url{https://data.bls.gov/timeseries/lns14000000}};
    \item \textbf{Gasoline}: ``US finished motor gasoline product supplied'' dataset\footnote{Publicly available at \url{https://www.eia.gov/dnav/pet/hist/LeafHandler.ashx?n=PET&s=wgfupus2&f=W}};
    \item \textbf{Sunspots}: ``Monthly numbers of sunspots, as from the World Data Center'' dataset\footnote{Publicly available at \url{http://www.sidc.be/silso/datafiles}};
    \item \textbf{Electricity}: ``Half-hourly electricity demand in England'' dataset\cite{Taylor2003}.

\end{itemize}

%Our first case study uses a relatively small dataset where classical SV validation method may not be optimal as the data are scarce. We refer to this ``monthly unemployment rate in US from 1948 to 1977'' dataset\footnote{Publicly available at \url{https://data.bls.gov/timeseries/lns14000000}} as \textbf{Labour}. To test a scenario where more data are available we used ``US finished motor gasoline product supplied'' dataset\footnote{Publicly available at \url{https://www.eia.gov/dnav/pet/hist/LeafHandler.ashx?n=PET&s=wgfupus2&f=W}}, which we refer to as \textbf{Gasoline}. Moreover, the ``half-hourly electricity demand in England'' dataset is used. We refer to this dataset as ``Electricity'' \cite{Taylor2003}. Lastly, a classical ``Monthly numbers of sunspots, as from the World Data Center''\footnote{Publicly available at \url{http://www.sidc.be/silso/datafiles}} dataset is used, which we refer to as ``sunspots''.

Lengths of these datasets and testing, validation split parameters are presented in Table \ref{tab:generative_datasets}. ``Min. ratio'' here is the percentage of the whole data (excluding testing) used as the minimal training length in AV, or the whole training length in FV (``Min.'' in Figure \ref{fig:validation_types}). 
%E.g., the classical SV validation method may not be optimal for Labour data, as it is so scarce: only 360 samples in total. 

\begin{table}[h]                           
 \centering
    \caption{Datasets and validation setup parameters. }
    \begin{tabular*}{\textwidth}{@{\extracolsep{\fill}} l r r r r}
    \toprule
     \tableHeader{Dataset} & \tableHeader{Samples $T$} & \tableHeader{Valid, test samples} & \tableHeader{Folds, steps $k$} & \tableHeader{Min. ratio}\\
    \midrule
     Labour & 360 & 10 & 34 & 50\% \\
    %  \midrule
     Gasoline & 1355 & 67 & 18 & 50\% \\ 
    %  \midrule 
      Sunspots & 3177 & 200 & 10 & 50\% \\ 
      Electricity & 4033 & 200 & 18 & 50\% \\ 
     \bottomrule  
    \end{tabular*}
    \label{tab:generative_datasets}                            
\end{table}

For $k$-fold CV, initial transient length and $k$ are chosen in such a way, that the $k$ folds would have the same size as testing. For $k$-step validation variants the overlapping validation block used also have the same length as the testing range. 

We use a grid search to find the best ESN hyper-parameters. Reservoir size $\n{x} = 50$ was chosen, and candidates of leaking rate $\alpha \in \{ 0.1, 0.2, 0.3,$ $ ...,  1\}$, spectral radius $\rho \in \{0.1, 0.2, 0.3, ..., 1.5\}$ and regularization degree $\beta \in \{ 0, 10^{-9}, 10^{-8}, ..., 1 \}$ were evaluated following \cite{Lukosevicius12a}.

\begin{table}[h]                           
 \centering
    \caption{Validation and testing NRMSEs on generative datasets}
        \begin{tabular*}{\textwidth}{ @{\extracolsep{\fill}} l l r r r r r r r r  }
    
    \toprule
     \multicolumn{2}{c} {\tableHeader{Method}} & \multicolumn{2}{c}{\tableHeader{Labour}}  &  \multicolumn{2}{c}{\tableHeader{Gasoline}} &       \multicolumn{2}{c}{\tableHeader{Sunspots}} & \multicolumn{2}{c}{\tableHeader{Electricity} } 
     \\
                 \cmidrule(lr){1-2} \cmidrule(lr){3-4} \cmidrule(lr){5-6} \cmidrule(lr){7-8} \cmidrule(lr){9-10}  
    \multicolumn{1}{l} {\tableHeader{Validation}} &    \multicolumn{1}{l} {\tableHeader{Final}} & \textbf{Valid} & \textbf{Test} & \textbf{Valid}& \textbf{Test}& \textbf{Valid}& \textbf{Test} & \textbf{Valid}& \textbf{Test} \\
    \midrule
     \multirow{2}{*}{SV} & As is & \multirow{2}{*}{1.034} & 1.927 & \multirow{2}{*}{0.891} & 0.881 & \multirow{2}{*}{0.749} & 0.784 &  \multirow{2}{*}{0.623} & 0.860  \\ 
     &  Retrained &     & 1.957 &  & 1.132 &   & 0.755 &  & 0.835  \\ 
        \midrule
     \multirow{3}{*}{$k$-fold CV} & Averaged & \multirow{3}{*}{2.009} & 1.835 & \multirow{3}{*}{1.000} & 0.914 & \multirow{3}{*}{1.060} &0.924 & \multirow{3}{*}{0.834} & 0.990 \\ 
     & Retrained & & 1.833 &  & 0.913 & {} & {0.970} &  & 1.006  \\ 
          &Best &  & 1.838 &  & 0.901 & {} & {1.008} &  & 0.995          \\    
    \midrule 
     \multirow{3}{*}{$k$-step AV} &Averaged  & \multirow{3}{*}{1.927}& 4.469 & \multirow{3}{*}{1.040} & 0.867  & \multirow{3}{*}{0.703} & 0.835 &  \multirow{3}{*}{0.812} & 0.829       \\ 
         &Retrained & &1.171 & & 0.962  & {} & {0.742} &    & 1.006         \\ 
         &Best &  &4.546 &  & \textbf{0.829}  & {} & {0.855} &  & 0.820          \\ 
            \midrule 
     \multirow{3}{*}{$k$-step FV}&Averaged  & \multirow{3}{*}{2.188} & 3.413 & \multirow{3}{*}{1.065} & 0.925   &  \multirow{3}{*}{0.726} & 0.640 & \multirow{3}{*}{0.783} & \textbf{0.733}  \\ 
         &Retrained &  & \textbf{0.681} &  & 0.949 & {} & {\textbf{0.612}} & & 1.006 \\ 
         &Best &  & 2.799 &  & 0.894 & {} & {0.649} &  & 0.769\\ 
    \bottomrule
    \end{tabular*}
    \label{tab:generative_results}                            
\end{table}

The experiment results are presented in Table \ref{tab:generative_results}. We can see that in all the experiments either FV or AV find the hyper-parameters producing best generalizing models, it is never SV or CV. The relative underperformance of CV can probably be explained by the nature of the non-stationary of the temporal data. The generating processes likely have a one-directional ``drift'', thus validation schemes FV and AV, that select models capable of predicting sequences directly following the training ones, win. They also win over SV, as this selection is validated over $k$ splits instead of just one.

The main bottleneck with the Labour dataset is its scarcity: only 360 samples in total. We see that SV overfits the hyper-parameters on the single split; CV gets a better estimate; and AV and FV fail in averaged and best modes, as these use the scarce data inefficiently, but produce the very best overall results when retrained. When having more data the benefits of retraining are less evident.

%Also, as the training set of Labour dataset was relatively small, signs of overfitting arise??.

\begin{figure*}
\makebox[\linewidth][c]{%
\centering
\subfigure[SV Retrained]{\label{fig:a}\includegraphics[width=0.5\textwidth]{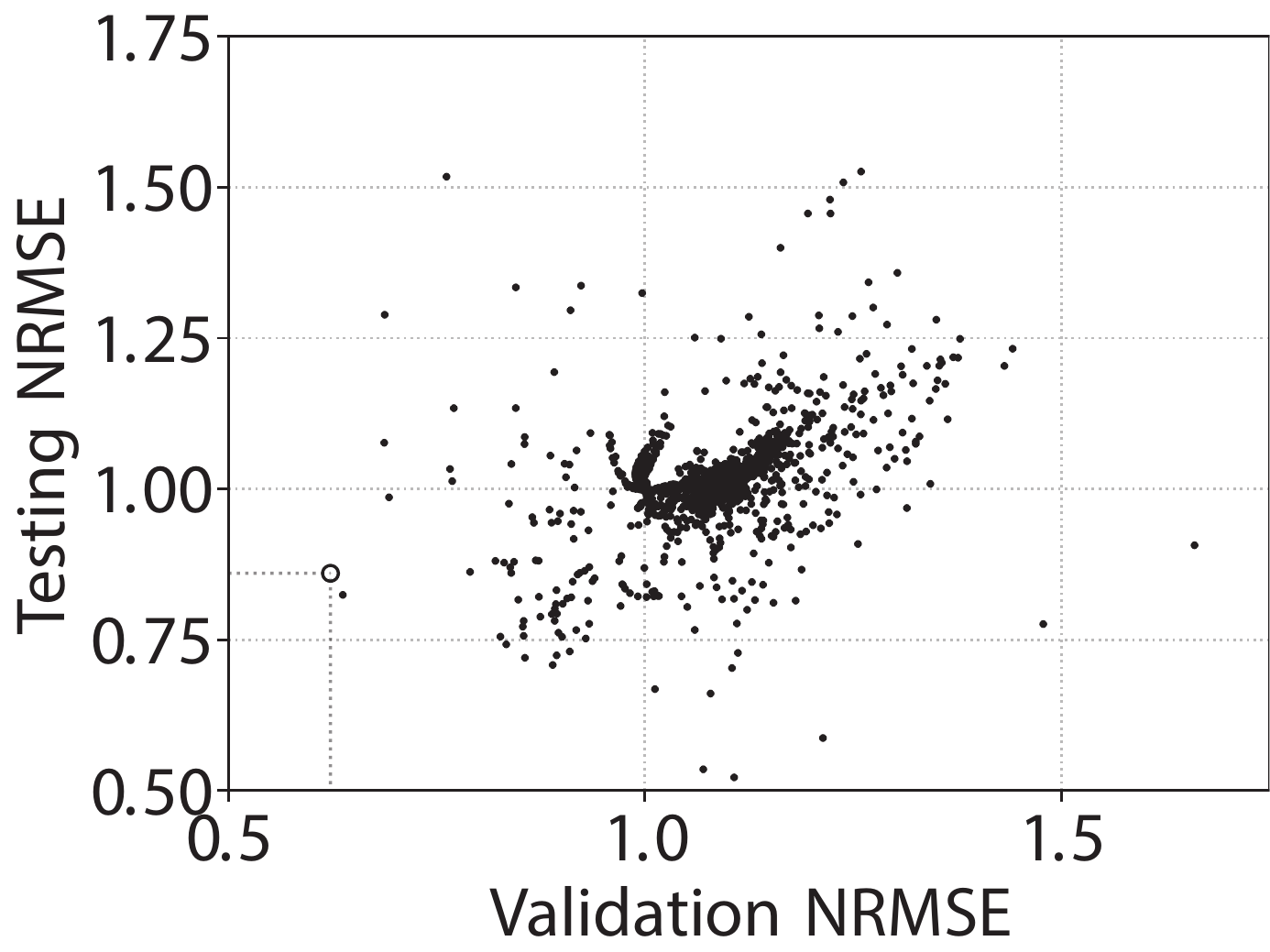}}%
\subfigure[$k$-step FV Averaged]{\label{fig:b}\includegraphics[width=0.5\textwidth]{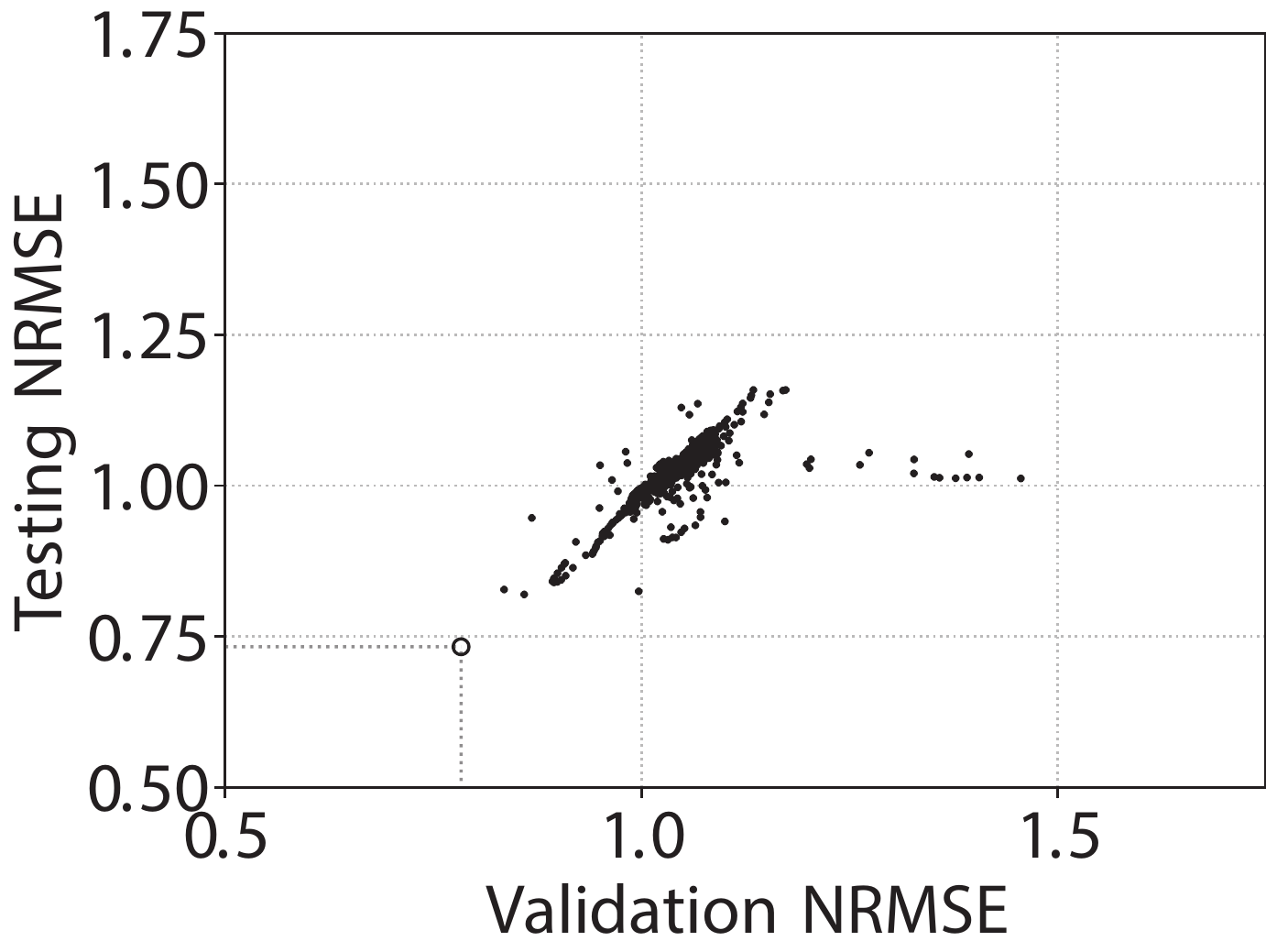}}%
}
\caption{Results of grid search on Electricity dataset. Every point corresponds to one combination of hyper-parameters.}
\label{fig:val_test}
\end{figure*}

Validation vs. testing errors of two validation schemes on the Electricity dataset are illustrated in Figure \ref{fig:val_test}. We can see that while there exist outliers with very good testing performance in Figure \ref{fig:a}, they would not be picked up by the validation. In fact, $k$-step FV validation errors for these hyper-parameter sets were so bad, that they went off-scale in Figure \ref{fig:b}. This indicates that the lucky outliers were particular to the testing data and most likely would not do well on other. On the other hand, $k$-step FV gives a much better overall correlation between validation and testing errors and a much better solution based on validation is picked in Figure \ref{fig:b} (the small circles).

\subsection{Time Series Classification}\label{classificationExperiments}
%Japanese vowels?...
To evaluate the validation methods on classification tasks we use a classical Japanese Vowels dataset\footnote{Publicly available at \url{https://archive.ics.uci.edu/ml/datasets/Japanese+Vowels} }. This benchmark comes as 270 training samples and 370 testing samples, where each sample consists of varying length 12 LPC cepstrum coefficients taken from nine male speakers. The goal of this task is to classify a speaker based on his pronunciation of vowel /ae/. In the training set, there are 30 samples for each user, while in the testing set, this number varies from 29 to 88.

We note that 0 test set misclassifications have been previously achieved with ESNs as reported in \cite{JaegerEtAl07si}. Therefore we shift our efforts to achieve better results on models that reportedly have been performing sub-optimally. %No grid search was used? 
It has been reported that models that only store the last state vector $\x(n)$ for every speaker have at best been able to achieve 8 test misclassifications. We replicate the model used in \cite{JaegerEtAl07si}, and set up a grid search as described in Section \ref{generativeExperiments}. We run an $18$-fold cross-validation, and use a validation length of 15 instances. As the dataset consists of a permutable set of sequences (the order does not matter), we only test the classical SV and the standard $k$-fold CV. We run the experiments 5 times having different random initializations of the ESN weights. For each of the 5 experiments we run grid search separately and report aggregated results. 

We also do individual regularization for each validation split. When doing cross-validation, each split on each regularization degree candidate is evaluated. $\W{out}$s of each split with their individual best regularization degrees are averaged. We refer to this variation as ``IReg Averaged''. In another option, the model was retrained on the whole data using the average of the best regularization degrees. We refer to this variation as ``IReg Retrained''. 
%\arnas{Nelogiskai padaryt, nes SV irgi naudoja tik 15 samplu kaip valid ..}
%\begin{table}[H]                           
 %\centering
    %\caption{Grid search setup}
    %\begin{tabular}{c c c c}
    %\toprule
     %$\n{R}$ & \tableHeader{$\alpha \in$} & \tableHeader{$\rho \in$} & reg $\in$ \\
    %\midrule
     %50 & $\{0.1i | 1 \le i \le 10 , i \in \mathbb{N} \}$ &$\{0.1i | 1 \le i \le 10 , i \in \mathbb{N} \}$ & $\{10^{-i} | 1 \le i \le 10 , i \in \mathbb{N} \}$    \\ 
    %\bottomrule
    %\end{tabular}
    %\label{tab:jap_grid_search}                            
%\end{table}
\begin{table}[t] 
    \centering
    \caption{Average results on Japanese Wovels task}
    \begin{tabular*}{\textwidth}{@{\extracolsep{\fill}} l l r r r}
    \toprule
     \tableHeader{Method} & \tableHeader{Final} & \tableHeader{Validation error} & \tableHeader{Test error} & \tableHeader{Misclasifications} \\
    \midrule
     \multirow{2}{*}{SV} &  As is & \multirow{2}{*}{0.504 $\pm$ 0.017}  & 0.491 $\pm$ 0.005 & 5.0 $\pm$ 1.5 \\ 
      & Retrained &  & 0.486 $\pm$ 0.003 & 4.8 $\pm$ 1.6 \\ 
        \midrule
         \multirow{2}{*}{CV}&  Averaged  & \multirow{2}{*}{0.493 $\pm$ 0.004} & 0.472 $\pm$ 0.008 & 4.2 $\pm$ 1.8 \\
         & Retrained &  & $  0.468 $ $\pm$ 0.006 & 4.4 $\pm$ 2.1 \\
        \midrule
         \multirow{2}{*}{CV IReg} & Averaged & \multirow{2}{*}{0.489 $\pm$ 0.004} & \textbf{0.468} $\pm$ 0.009 & 4.4 $\pm$ 1.2 \\
         {} & Retrained&  & $ 0.470 $ $\pm$  0.008 & \textbf{3.8} $\pm$ 1.8 \\
        \bottomrule
    \end{tabular*}
    \label{tab:jap_grid_search}                                

\end{table}

The NRMSE errors, misclassifications and their standard deviations are presented in Table \ref{tab:jap_grid_search}. We see that all the CV variations outperform all the SV variations. We also see that the individual regularization ``IReg'' further slightly improves both validation and testing errors, as well as misclassification, which is not surprising.
%We can see that both the validation and the testing errors are the lowest when validating by the ``CV IReg'' method. Because each fold has more degrees of freedom while applying multiple regularization degrees and picking the best, there is no surprise that the validation error is the lowest. 
In all validation schemes, except the ``CV IReg Averaged'', there was at least one model produced that was able to achieve 2 test misclassifications.

%\arnas{Gal tokia struktura informatyviau?}
%\subsection{Datasets}

%\subsection{Results}

\section{Discussion}\label{discussion}

In this contribution we have proposed and motivated different cross-validation schemes for ESNs, have introduced a space- and time-efficient algorithm for doing this, and empirically investigated their effects on several real-world datasets.

The component that dominates the time complexity of the already quite fast ESN training remains constant (does not scale up with $k$) in our proposed method of doing $k$-fold cross-validation. The component that does scale linearly with $k$ starts dominating only in some not very common situations, in particular when $k$ is very large. Thus in typical situations $k$-fold cross-validation of ESNs can be done for virtually the same time complexity as a simple single validation. The time savings are also less evident when the data are short, but then they are also less pertinent. The methods can also have the same space complexity as a simple single validation. 

This further sets apart the speed of ESN training from error backpropagation based recurrent neural network training methods, where cross-vali\-da\-tion could also be used in principle.  

We have demonstrated the proposed validation schemes for classical ESNs, but they apply to other reservoir types as well with the same time and space complexity savings. Namely, the time and space complexity of running the reservoirs is added to the ones of training instead of multiplying them by $k$ in a $k$-fold/step cross-validation. The time complexity savings on reservoir running could also hold for other readout types. 

We have also empirically investigated the benefits of the proposed validation schemes for ESNs on several different datasets. There is no single winner among the validation schemes. The results highly depend on the nature of the data. Overall our experiments show that typically cross-validation predicts testing errors more accurately and produces more robust results. It also can use scarce data more sparingly. AV and FV validation schemes can apparently select for good ``forward-predicting'' models when the generating process is ``evolving'' and not quite stationary. SV can also have its benefits, as it can be seen as the last fold of AV. For stationary and ample, well-representative data cross-validation might not be crucial. 

How the final trained model is produced from the cross-validation results is also very important. 

%how stationary the data are. If it is not quite stationary, learning and validating tasks with it is intrinsically difficult. This is especially the case with generative tasks where the errors tend to explode. The proposed validation methods prove to be beneficial in any case.

%\arnas{Pamineti, kad advantage pries backprop-trainnable RNN?} Because of backpropogation through time and echo state property, and one-shot learning.

%Results depend on data stationarity.

%Validating generative mode is intrinsically difficult.

\section*{Acknowledgments}

This research was supported by the Research, Development and Innovation Fund of Kaunas University of Technology (grant No. PP-91K/19).

%\newpage

\bibliography{ML2}

\begin{thebibliography}{10}

\bibitem{Jaeger01a}
Herbert Jaeger.
\newblock The ``echo state'' approach to analysing and training recurrent
  neural networks.
\newblock Technical Report GMD Report 148, German National Research Center for
  Information Technology, 2001.

\bibitem{JaegerHaas04}
Herbert Jaeger and Harald Haas.
\newblock Harnessing nonlinearity: predicting chaotic systems and saving energy
  in wireless communication.
\newblock {\em Science}, 304(5667):78--80, 2004.

\bibitem{Jaeger07scholarpedia}
Herbert Jaeger.
\newblock Echo state network.
\newblock {\em Scholarpedia}, 2(9):2330, 2007.

\bibitem{LukoseviciusJaeger09}
Mantas Luko{\v{s}}evi{\v{c}}ius and Herbert Jaeger.
\newblock Reservoir computing approaches to recurrent neural network training.
\newblock {\em Computer Science Review}, 3(3):127--149, August 2009.

\bibitem{Stone74}
Mervyn Stone.
\newblock Cross-validatory choice and assessment of statistical predictions.
\newblock {\em Journal of the Royal Statistical Society: Series B
  (Methodological)}, 36(2):111--133, 1974.

\bibitem{Lukosevicius12a}
Mantas Luko{\v{s}}evi{\v{c}}ius.
\newblock A practical guide to applying echo state networks.
\newblock In Gr{\'{e}}goire Montavon, Genevi{\`{e}}ve~B. Orr, and Klaus-Robert
  M{\"{u}}ller, editors, {\em Neural Networks: Tricks of the Trade, 2nd
  Edition}, volume 7700 of {\em LNCS}, pages 659--686. Springer, 2012.

\bibitem{YildizEtAl12}
Izzet~B. Yildiz, Herbert Jaeger, and Stefan~J. Kiebel.
\newblock Re-visiting the echo state property.
\newblock {\em Neural Networks}, 35:1 -- 9, 2012.

\bibitem{BergmeirEtAl18}
Christoph Bergmeir, Rob~J Hyndman, and Bonsoo Koo.
\newblock A note on the validity of cross-validation for evaluating
  autoregressive time series prediction.
\newblock {\em Computational Statistics \& Data Analysis}, 120:70--83, 2018.

\bibitem{Taylor2003}
J~W Taylor.
\newblock Short-term electricity demand forecasting using double seasonal
  exponential smoothing.
\newblock {\em Journal of the Operational Research Society}, 54(8):799--805,
  aug 2003.

\bibitem{JaegerEtAl07si}
Herbert Jaeger, Mantas Luko\v{s}evi\v{c}ius, Dan Popovici, and Udo Siewert.
\newblock Optimization and applications of echo state networks with
  leaky-integrator neurons.
\newblock {\em Neural Networks}, 20(3):335--352, 2007.

\end{thebibliography}
\bibliographystyle{unsrt}

\end{document}